\documentclass[12pt]{article}
\pdfoutput=1
\usepackage{graphicx,pict2e,latexsym,xcolor}
\usepackage[square,numbers]{natbib}
\newtheorem{defn}{Definition}
\newtheorem{theorem}{Theorem}
\newtheorem{lemma}{Lemma}

\def\G{\mathbf G}

\def\L{\mathbf L}

\def\I{\mathbf I}
\def\U{\mathbf U}

\def\A{\mathbf A}
\def\X{\mathbf X}
\def\W{\mathbf W}

\newcommand{\iid}{\stackrel{iid}{\sim}}

\begin{document}
\bibliographystyle{plainnat}
\pagestyle{plain}

\title{\Large \bf Learning 1-Dimensional Submanifolds \\ 
for Subsequent Inference on \\ Random Dot Product Graphs}

\author{Michael W. Trosset\thanks{Department of Statistics, Indiana University.  
E-mail: {\tt mtrosset@indiana.edu}} 
\and
Mingyue Gao\thanks{Current affiliation: Financial Industry Regulatory Authority (FINRA); work performed while a Ph.D. student at Johns Hopkins University.}
\and
Minh Tang\thanks{Department of Statistics, North Carolina State  University. 
E-mail: {\tt mtang8@ncsu.edu}} 
\and
Carey E. Priebe\thanks{Department of Applied Mathematics \& Statistics, Johns Hopkins University. 
E-mail: {\tt cep@jhu.edu}}
}

\date{\today}

\maketitle

\newpage

\begin{abstract}
A random dot product graph (RDPG) is a generative model for networks in which vertices correspond to positions in a latent Euclidean space and edge probabilities are determined by the dot products of the latent positions.  We consider RDPGs for which the latent positions are randomly sampled from an unknown $1$-dimensional submanifold of the latent space.  In principle, restricted inference, i.e., procedures that exploit the structure of the submanifold, should be more effective than unrestricted inference; however, it is not clear how to conduct restricted inference when the submanifold is unknown.  We submit that techniques for manifold learning can be used to learn the unknown submanifold well enough to realize benefit from restricted inference.
To illustrate, we test $1$- and $2$-sample hypotheses about the Fr\'{e}chet means of small communities of vertices, using the complete set of vertices to infer latent structure.  We propose test statistics that deploy the Isomap procedure for manifold learning, using shortest path distances on neighborhood graphs constructed from estimated latent positions to estimate arc lengths on the unknown $1$-dimensional submanifold.  Unlike conventional applications of Isomap, the estimated latent positions do not lie on the submanifold of interest.  We extend existing convergence results for Isomap to this setting and use them to demonstrate that, as the number of auxiliary vertices increases, the power of our test converges to the power of the corresponding test when the submanifold is known.  Finally, we apply our methods to an inference problem that arises in studying the connectome of the {\em Drosophila}\/ larval mushroom body.  The univariate learnt manifold test rejects ($p<0.05$),
while the multivariate ambient space test does not ($p\gg0.05$),
illustrating the value of identifying and exploiting low-dimensional structure for subsequent inference.
\end{abstract}


\newpage

\tableofcontents

\newpage


\section{Introduction}
\label{intro}

Statistical inference requires specification of a probability model from which observed data are generated.  Accordingly, network scientists have proposed a variety of probability models for graphs.     A simple example of such a model is an Erd\"{o}s-R\'{e}nyi model, in which a finite set of vertices is fixed and edges are established by independent Bernoulli trials with a fixed success probability.  Stochastic blockmodels \cite{holland&etal:1983,anderson&etal:1992} generalize the Erd\"{o}s-R\'{e}nyi model; in turn, random dot product graphs \cite{young&scheinerman:2007} generalize stochastic blockmodels.  
There now exists a considerable body of theory and methodology for statistical inference on random dot product graphs; see, for example, \cite{AA&etal:2018} and the references therein.

Random dot product graphs also exemplify the latent space approach to network analysis considered in \cite{hoff&etal:2002}.
Recently, Athreya and collaborators \cite{LSM:2018} studied the special case of a random dot product graph whose latent positions lie on a curve.  This is the case that concerns us herein, except we suppose that the curve of interest is completely unknown.  We submit that procedures that exploit the curve's structure should be more effective than procedures that do not.  The challenge is how to exploit that structure when it is unknown.  The methods that we propose can be extended to the case of generalized random dot product graphs, as defined in \cite{GRDPG:2017}, but the notation is more cumbersome.

To fix ideas, we study $1$- and $2$-sample testing problems.  For $1$-sample problems, we suppose that a small community of vertices correspond to a particular latent position; for $2$-sample problems, we suppose that two small communities of vertices correspond to two particular latent positions.  In both cases, we also assume
that a large number of additional vertices correspond to randomly generated latent positions.  Assuming that the latent positions lie on a curve, tests that restrict alternatives to the curve should be more powerful than tests that do not.  We use the additional vertices to learn the curve well enough to realize gains from restricted inference.

To learn submanifolds of latent positions, we apply the popular manifold learning procedure Isomap \cite{isomap:2000} to a set of estimated latent positions.  Because these estimates need not lie on the manifold of interest, the traditional analysis of Isomap convergence \cite{Bernstein&etal:2000} does not apply.  To address this difficulty, we present a novel analysis of Isomap convergence in the presence of noise.

In what follows, Section~\ref{rdpg} contains a succinct exposition of random dot product graphs.  Sections~\ref{test1} and~\ref{test2} describe $1$- and $2$-sample testing problems.  Tests are proposed for three situations: an unrestricted test in the ambient space, a restricted test when the $1$-dimensional submanifold of latent positions is known, and a restricted test when the submanifold is not known.  
Section~\ref{isomap} contains our convergence analysis of Isomap, which may be of independent interest.  Section~\ref{power} compares the power of the two restricted tests.  Section~\ref{app} applies our methods to the study of the {\em Drosophila}\/ larval mushroom body connectome.   Section~\ref{discuss} concludes.

\section{Random Dot Product Graphs}
\label{rdpg}

A {\em graph}\/ is an ordered pair ${\mathcal G}=(V,E)$, where $V=\{1,\ldots,n\}$ is the {\em vertex set}\/ and $E \subset V \times V$ is the {\em edge set}.  There is an edge between vertices $i$ and $j$ if and only if $(i,j) \in E$.  Alternatively, the $n \times n$ {\em adjacency matrix}\/ $\A$ of ${\mathcal G}$ is the 0-1 matrix defined by $A_{ij}=1$ if and only if $(i,j) \in E$.
The graph ${\mathcal G}$ is undirected if $\A$ is symmetric and simple if $\A$ is hollow, i.e., if each $A_{ii}=0$.  A {\em random graph}\/ is a probability model for generating graphs, or (equivalently) adjacency matrices.

Latent space models for random graphs assume that vertices correspond to points in some space and that the probability of observing an edge between vertices $i$ and $j$ is a {\em link function}\/ of their corresponding latent positions.  The random dot product graph (RDPG) model assumes that the link function is the Euclidean inner product.  
Our exposition follows \cite{AA&etal:2018}.

\begin{defn}[RDPG]
Fix $X_1,\ldots,X_n \in \Re^k$, then form the $n \times k$ latent position matrix
\[
\X = \left[ \begin{array}{c|c|c} X_1 & \cdots & X_n \end{array} \right]^\top.
\]
Suppose that $\A$ is a symmetric hollow adjacency matrix whose above-diagonal entries are independent Bernoulli trials with success probabilities $P(A_{ij}=1)=X_i^\top X_j$.  We then write $\A \sim \mbox{RDPG}(\X)$ and say that $\A$ is the adjacency matrix of a random dot product graph with latent positions $X_1,\ldots,X_n$.
\end{defn}

Like \cite{LSM:2018},
the present manuscript is concerned with drawing inferences about RDPGs when the latent positions are restricted to lie on a $1$-dimensional submanifold ${\mathcal H} \subset \Re^k$.  

\subparagraph{Example 1}
Define $\psi : [0,1] \rightarrow \Re^3$ by $\psi(\tau) = (\tau^2,2 \tau (1-\tau), (1-\tau)^2)$ and suppose that the latent positions lie on the curve ${\mathcal H} = \psi([0,1])$.
Viewed as a subset of the unit simplex in $\Re^3$, ${\mathcal H}$ indexes the trinomial distributions in Hardy-Weinberg equilibrium, an important phenomenon in statistical genetics.  In the present context, ${\mathcal H}$ provides a convenient example of a special type of RDPG, a $1$-dimensional {\em latent structure model}, formally defined in \citep{LSM:2018}.
\hfill $\Box$

\bigskip

Notice that
the observed above-diagonal entries of $\A$ are unbiased estimates of the edge probabilities $P(A_{ij}=1)$  of the $\mbox{RDPG}(\X)$ from which $\A$ was drawn.  Hence, one can infer a plausible set of latent positions by constructing a set of points $\hat{X}_1,\ldots,\hat{X}_n \in \Re^r$ whose pairwise inner products approximate the above-diagonal entries of $\A$.  This is the premise of {\em adjacency spectral embedding}.

\begin{defn}[adjacency spectral embedding]
Let $\lambda_1 \geq \cdots \geq \lambda_r$ denote the $r$ largest eigenvalues of adjacency matrix $\A$ and let $u_1,\ldots,u_r$ denote corresponding eigenvectors.  Set $\sigma_i^2 = \max(\lambda_i,0)$.
The {\em adjacency spectral embedding}\/ (ASE) of $\A$ in $\Re^r$ is
$\hat{\X} = \U_r {\mathbf S}_r$, where $\U_r$ is the $n \times r$ matrix whose columns are $u_1,\ldots,u_r$ and ${\mathbf S}_r$ is the $r \times r$ diagonal matrix whose diagonal entries are $\sigma_1,\ldots,\sigma_r$.
\end{defn}

The edge probabilities of an RDPG depend on its latent positions only through their pairwise inner products.  If $\W$ is any $k \times k$ orthogonal matrix, then $(\X\W)(\X\W)^\top = \X\W\W^\top \X^\top = \X\X^\top$; hence, the latent positions $\X$ and $\X\W$ have the same edge probabilities, i.e., the latent positions are not identifiable.
Nevertheless, they can be consistently estimated in the following sense.
\begin{theorem}[\cite{VL&etal:2017,2toInf:2019,LSM:2018}]
Let
\[
\X_n = \left[ \begin{array}{c|c|c} X_{1} & \cdots & X_{n} \end{array} \right]^\top,
\]
with $\mbox{rank}(\X_n)=k$ for all sufficiently large $n$.
Set
$g_n = \max_i \sum_j X_{i}^\top X_{j}$,
the maximum expected degree of $\A_n$, which increases more rapidly than $\log^2 n$ as $n \rightarrow \infty$.
Suppose that
$\A_n|\X_n \sim \mbox{RDPG}(\X_n)$ and
let $\hat{X}_{n,1},\ldots,\hat{X}_{n,n}$ denote the ASE of $\A_n$ in $\Re^k$.
Then there exists $C>0$ and a sequence of $k \times k$ orthogonal matrices $\W_n$ such that
\[
\lim_{n \rightarrow \infty} P \left(
\max_{i=1,\ldots,n} \left\|
\W_n \hat{X}_{n,i} - X_{i} 
\right\| \leq C \left( k/g_n \right)^{1/2} \log^2 n 
\right) = 1.
\]
\label{thm:2inf}
\end{theorem}
Notice that Theorem~\ref{thm:2inf} guarantees {\em uniform convergence}\/ of the estimated latent positions to the true latent positions, a property that is crucial for our results.

If the latent positions are independently and identically sampled from a suitable probability distribution, then the estimates of the latent positions are asymptotically normal in the following sense.

\begin{defn}[inner product distribution]
A probability distribution $F$ with support ${\mathcal X} \subset \Re^k$ is a $k$-dimensional inner product distribution if and only if $x,y \in {\mathcal X}$ entails $x^\top y \in [0,1]$.
\end{defn}

\begin{defn}[RDPG with iid latent positions]
Suppose that $F$ is an inner product distribution and $X_1,\ldots,X_n \iid F$.
If $\A|\X \sim \mbox{RDPG}(\X)$, then we write $(\A,\X) \sim \mbox{RDPG}(F)$ and say that $\A$ is the adjacency matrix of a random dot product graph with random latent positions $X_1,\ldots,X_n$.
\end{defn}

\begin{theorem}[\cite{AA&etal:2016}]
Suppose that $F$ is an inner product distribution with support ${\mathcal X} \subset \Re^k$.  
For $X_1 \sim F$, set $\G = E_F X_1X_1^\top$ and
\[
{\mathbf \Sigma}(x) = \G^{-1}
E_F \left[ \left(
x^\top X_1 - \left( x^\top X_1 \right)^2 \right) X_1X_1^\top 
\right] \G^{-1}.
\]
Suppose that $(\A_n,\X_n) \sim \mbox{RDPG}(F)$ and let $\hat{\X}_n$ denote the ASE of $\A_n$.
Then there exists a sequence of $k \times k$ orthogonal matrices $\W_n$ such that, for any fixed $i$ and every $z \in \Re^k$,
\[
\lim_{n \rightarrow \infty} P \left( n^{1/2} 
\left[ \mbox{row $i$ of } \W_n \hat{\X}_n - \X_n \right] \leq z \right) = \int_{\mathcal X} \Phi \left( z, {\mathbf \Sigma}(x) \right) F(dx),
\]
where $\Phi(z,{\mathbf \Sigma}(x))$ is the cumulative distribution function of a multivariate normal distribution with mean vector zero and covariance matrix ${\mathbf \Sigma}(x)$.
\label{thm:CLT}
\end{theorem}

\section{One-Sample Tests}
\label{test1}

Suppose that $\psi : [0,1] \rightarrow \Re^k$ and that ${\mathcal H} = \psi([0,1])$ is a $1$-dimensional submanifold of $\Re^k$.
Formal statement of the inference problem that we investigate requires the concept of the Fr\'{e}chet mean of a probability distribution $h^*$ on ${\mathcal H}$.  The following is a special case of Definition~2.1 in \cite{bhattacharya&patrangenaru:2003}.
\begin{defn}[Fr\'{e}chet mean]
The {\em Fr\'{e}chet mean set}\/ of $h^*$ is the set of all minimizers of the map $\mbox{\rm Fr} : {\mathcal H} \rightarrow \Re$ defined by
\[
\mbox{\rm Fr}(p) = \int_{\mathcal H} \left[
d_M (p,x) \right]^2 h^*(dx).
\]
If a unique minimizer $\mu_{\mbox{\rm \tiny Fr}}$ exists, then it is the {\em Fr\'{e}chet mean}\/ of $h^*$.
\end{defn}

The sample Fr\'{e}chet mean of $x_1,\ldots,x_s \in {\mathcal H}$ is the Fr\'{e}chet mean of the empirical distribution of $x_1,\ldots,x_s$.  
Theorem~2.3 in \cite{bhattacharya&patrangenaru:2003} 
establishes that the sample Fr\'{e}chet mean is a strongly consistent estimator of the Fr\'{e}chet mean.
In general, the sample Fr\'{e}chet mean is difficult to compute.
In the present setting, however,
suppose that $\gamma : [0,L] \rightarrow \Re^k$ parametrizes 
${\mathcal H}$ by arc length and that $x_i = \gamma(t_i)$.
Writing $p=\gamma(t)$, we obtain
\[
\mbox{Fr}(t) = \frac{1}{s} \sum_{i=1}^s \left[ \int_t^{t_i} 1 \, dt \right]^2 = \frac{1}{s} \sum_{i=1}^s \left[ t_i-t \right]^2,
\]
which is minimized by $t=\bar{t}$, the sample mean of $t_1,\ldots,t_s$.
The sample Fr\'{e}chet mean is therefore $p=\gamma(\bar{t})$.

Now suppose that $\A$ is an $n \times n$ adjacency matrix generated by an RDPG whose latent positions lie in ${\mathcal H}$.
Assume that the latent positions are of two types:
\begin{enumerate}

\item  A small number ($s$) of latent positions $p_1,\ldots,p_s$ with a fixed location $p^* = \psi(\tau^*)$, or sampled from a small neighborhood thereof.  
These latent positions form a known community about which we hope to test the simple null hypothesis $H_0: p^* = p_0$ against the composite alternative hypothesis $H_1 : p^* \neq p_0$.  
Formally, let $h^*$ be a probability distribution on ${\mathcal H}$ with Fr\'{e}chet mean $p^*$ and assume that $p_1,\ldots,p_s \iid h^*$.

\item  A large number ($m$) of auxiliary latent positions $p_{s+1},\ldots,p_{s+m}$ that provide information about the structure of ${\mathcal H}$.  Let $p_i = \psi(\tau_i)$.  We assume that $\tau_{s+1},\ldots,\tau_{s+m} \iid \mu$, where $\mu$ is a strictly positive probability density function on $[0,1]$.
Our numerical experiments set $\mu = \mbox{Uniform}[0,1]$.

\end{enumerate}
Collectively, $n=s+m$, the $n \times k$ latent position matrix is
\[
\X = \left[ \begin{array}{c|c|c|c|c|c} p_1 & \cdots & p_s & p_{s+1} & \cdots & p_{s+m} \end{array} \right]^\top,
\]
and $\A \sim \mbox{RDPG}(\X)$.  Typically $m \gg s$.  In the spirit of \cite{mwt:lrt,mwt:ait}, we propose tests that attempt to exploit the structure of ${\mathcal H}$.

Having observed $\A \sim \mbox{RDPG}(\X)$, we first construct $\hat{X}_1,\ldots,\hat{X}_n \in \Re^r$,
then compute various test statistics that depend on $\hat{\X}$.  Two difficulties are immediately apparent.  First, if $k$ is unknown, then how is $r$ determined?  A common approach involves examining a scree plot of the singular values of the ASE and identifying an ``elbow'' by some means, either heuristically or automatically.  Alternatively, one might circumvent this difficulty by testing with multiple embeddings, obtaining a significance probability for each, choosing the smallest, and adjusting for multiple comparisons, as in \cite{mirkat:2015}.

Second, because the true latent positions are not identifiable, $\hat{\X}$ must be rotated to the representation in which the null hypothesis is specified.  Implicit in the specification $H_0: p^* = p_0$ is the assumption that one has access to information about the desired representation that allows one to determine the proper rotation.  Conceptually, one could align $\hat{X}_1,\ldots,\hat{X}_n$ with $p_1,\ldots,p_n$ by performing a Procrustes analysis, i.e.,
one would choose $\hat{\W} = \hat{\W}(\A,\X)$ to minimize 
\[
\sum_{i=1}^n \left\| \W\hat{X}_i-p_i \right\|_2^2
\] 
in the group of $k \times k$ orthogonal matrices.  In practice, of course, this calculation is impossible because $p_1,\ldots,p_n$ are unknown.  This difficulty disappears in the case of $2$-sample problems, as discussed in Section~\ref{test2}.   

Neither of these difficulties concern us here, as
our present interest lies in the possibility of exploiting the manifold structure of ${\mathcal H}$ to construct restricted tests.
To demonstrate the potential of restricted testing, we simply assume that both $r=k$ and the optimal rotation $\hat{\W}$ are known.

\subsection{Unrestricted Testing}
\label{unrestricted}

The vertices of interest are $1,\ldots,s$.
Let
\[
\bar{X}_s = \frac{1}{s} \sum_{i=1}^{s} \hat{\W} \hat{X}_i
\]
denote the centroid of their estimated positions after rotation.
For any symmetric positive definite $k \times k$ matrix $\L$,
define the unrestricted test statistic
\[
{T}_k^2(\A) = \left( \bar{X}_s - p_0 \right)^\top \L
\left( \bar{X}_s - p_0 \right).
\]
Critical regions of such tests are of the form
$\{ \A : {T}_k(\A) \geq c > 0 \}$.

If $\L=\I$, then
${T}_k(\A) = \| \bar{X}_s - p_0 \|_2$ is
the Euclidean distance in $\Re^k$ of the centroid from the hypothesized latent position.  
If $\L = {\mathbf \Sigma} \left( p_0 \right)^{-1}$, then Theorem~\ref{thm:CLT} suggests that
${T}_k(\A)$ is approximately the hypothesized Mahalanobis distance of the centroid from the hypothesized latent position.
Unless the inner product distribution $F$ is known, the covariance matrix ${\mathbf \Sigma} (p_0)$ cannot be computed; however, estimating it from the sample covariance matrix of $\hat{X}_1,\ldots,\hat{X}_s$ leads to Hotelling's $1$-sample $T^2$ test.

\subsection{Testing Restricted to the True Submanifold}
\label{restricted}

Suppose that $p_1,\ldots,p_n \in {\mathcal H} = \psi([0,1])$, where $\psi : [0,1] \rightarrow \Re^k$ is known, and define $\tau_0$ by $p_0=\psi(\tau_0)$. 
If $p_1,\ldots,p_s$ could be observed, then a natural test statistic would be the geodesic (arc length) distance in ${\mathcal H}$ from the hypothesized Fr\'{e}chet mean $p_0$ to the sample Fr\'{e}chet mean of $p_1,\ldots,p_s$.
As $p_1,\ldots,p_s$ are not observed, we replace each $p_i$ with a corresponding $\hat{p}_i \in {\mathcal H}$.

We proceed by minimum distance estimation \cite{basu&etal:2011}.
If $n$ is large, then the distribution of $\sqrt{n} \left[ \hat{\W} \hat{X}_i - p_i \right]$ will be well-approximated by a multivariate normal distribution with covariance matrix ${\mathbf \Sigma}(p)$.
This fact suggests estimating $p_i=\psi(\tau_i)$ by 
$\hat{p}_i=\psi(\hat{\tau}_i)$, where
$\hat{\tau}_i$, minimizes the objective function
\[
\mbox{MDE}(\tau) = \left[ \psi(\tau) - \hat{\W} \hat{X}_i \right]^\top
\left[ {\mathbf \Sigma} ( \psi(\tau) ) \right]^{-1}
\left[ \psi(\tau) - \hat{\W} \hat{X}_i \right].
\]
This approach has two potential drawbacks.  First, if the inner product distribution $F$ is unknown, then ${\mathbf \Sigma}(p)$ cannot be computed.  Second, even if $F$ is known, minimizing $\mbox{MDE}(\tau)$ may be difficult.  
To mitigate these difficulties,
we simplify $\mbox{MDE}(\tau)$ by replacing ${\mathbf \Sigma}(p)$ with a fixed symmetric positive definite $k \times k$ matrix $\L$.
 
Assuming that $\| \dot{\psi} (\tau) \|$ is bounded,
the restricted test statistic $T_1(\A)$ is the geodesic distance $d_M$ from the hypothesized Fr\'{e}chet mean $p_0$ to the sample Fr\'{e}chet mean of $\psi(\hat{\tau}_1),\ldots,\psi(\hat{\tau}_s)$. 
To compute $T_1(\A)$, first compute
\[
t_0 = \int_0^{\tau_0}
\left\| \dot{\psi}(\tau) \right\| \, d\tau, \quad
\hat{t}_i = \int_0^{\hat{\tau}_i} 
\left\| \dot{\psi}(\tau) \right\| \, d\tau, \quad \mbox{and} \enspace
\bar{t} = \frac{1}{s} \sum_{i=1}^s \hat{t}_i. 
\]
Then
\[
{T}_1 ( \A ) = d_M \left(
\mu_{\mbox{\rm \tiny Fr}} \left( \psi(\hat{\tau}_1),\ldots,\psi(\hat{\tau}_s) \right) , p_0 \right) =
\left| \bar{t}-t_0 \right| =
\left| \frac{1}{s} \sum_{i=1}^s
\int_{\tau_0}^{\hat{\tau}_i} \left\| \dot{\psi}(\tau) \right\| \, d\tau \right|.
\]
Critical regions of these tests are of the form
$\{ \A : {T}_1(\A) \geq c > 0 \}$.

If ${\mathcal H}=\psi([0,1])$ is not parametrized by arc length, then equal increments in $\tau$ may not correspond to equal increments in arc length.

\subparagraph{Example 1 (continued)}
If $\psi(\tau) = (\tau^2,2\tau(1-\tau),(1-\tau)^2)$, then
$\left\| \dot{\psi} (\tau) \right\|^2 = 8 \left( 3\tau^2-3\tau+1 \right) \leq 8$.  Suppose that the hypothesized Fr\'{e}chet mean $p_0 = \psi(0.3)$.  If the sample Fr\'{e}chet mean is
$\psi(0.55)$, then $\bar{T}_1 ( \A ) \doteq 0.375$.
If the sample Fr\'{e}chet mean is
$\psi(0.05)$, then $\bar{T}_1 ( \A ) \doteq 0.536$.
Although the parameter value $0.3$ lies midway between the parameter values $0.05$ and $0.55$, the point $\psi(0.3)$ does not lie midway along the arc between the points $\psi(0.05)$ and $\psi(0.55)$.
This phenomenon occurs because $\psi$ does not parametrize ${\mathcal H}$ at constant speed, i.e.,
equal increments of $\tau$ may not correspond to equal arc lengths of ${\mathcal H}$.  Notice that the choice of parametrization would not matter if we were concerned with one-sided alternatives, as the arc length distance of $\psi(\tau_0+\tau)$ from $\psi(\tau_0)$ is an increasing function of $|\tau|$.  
\hfill $\Box$

\subsection{Testing Restricted to a Learnt Submanifold}
\label{restricted.learnt}
Suppose that we wish to test $H_0: p^* = p_0$ but that we lack knowledge of ${\mathcal H}$.
The test statistic $T_1$ is based on the concept of arc length in ${\mathcal H}$; hence,
if arc lengths can be estimated directly from the $\hat{\W} \hat{X}_i$, then it may be possible to approximate $T_1$ without knowing ${\mathcal H}$.
In fact, the manifold learning procedure Isomap \cite{isomap:2000}, which approximates geodesic distances on a manifold with shortest path distances on a graph, does precisely that.
Isomap is described in Figure~\ref{fig:isomap}.

\begin{figure}[tb]
\begin{center}
\fbox{ 
\begin{minipage}{5in}
\vspace{1em}
Given: feature vectors $x_1,\ldots,x_m \in \Re^k$ and a target dimension $d$.
\begin{enumerate}

\item Construct a $\lambda$-neighborhood or $K$-nearest-neighbor graph of the observed feature vectors.
Weight edge $i \leftrightarrow j$ of the graph by $\| x_i-x_j \|$.

\item Compute the dissimilarity matrix $\Delta = [ \delta_{ij} ]$,
where $\delta_{ij}$ is the shortest path distance between vertices $i$
and $j$.  The key idea that underlies Isomap is that shortest
path distances on a locally connected graph approximate 
geodesic distances on an underlying manifold.

\item Embed $\Delta$ by classical multidimensional scaling (CMDS), obtaining $z_1,\ldots,z_m \in \Re^d$.  

\end{enumerate}
\vspace{1em}
\end{minipage}
}
\end{center}
\caption{Isomap, the manifold learning procedure proposed in \cite{isomap:2000}.  In step~3, we replace CMDS, embedding $\Delta$ by minimizing the raw stress criterion.}
\label{fig:isomap}
\end{figure}

As originally proposed, Isomap obtains a configuration $z_1,\ldots,z_m \in \Re^d$ from $\Delta = [ \delta_{ij} ]$, the matrix of pairwise shortest path distances, by classical multidimensional scaling (CMDS).  However, if one seeks to approximate the shortest path distances with Euclidean distances, then it is more natural to embed $\Delta$ by choosing
$z_1,\ldots,z_m \in \Re^d$
to minimize the {\em raw stress criterion},
\[
\sigma \left( z_1,\ldots,z_m \right) =
\frac{1}{2} \sum_{i,j=1}^m u_{ij} \left( \left\| z_i-z_j \right\| - \delta_{ij} \right)^2.
\]
This criterion is often minimized by repeated iterations of the {\em Guttman transformation}, described in \cite[Chapter 8]{borg&groenen:2005}.  At least when $u_{ij}=1$ and the configuration is initialized by CMDS, several iterations usually result in a nearly optimal embedding.

Traditionally, Isomap is deployed when $x_1,\ldots,x_n \in \Re^k$ lie on a $d$-dimensional data manifold.
In the present application, $x_1,\ldots,x_n$ are estimated latent positions.  The true latent positions lie on a $1$-dimensional manifold, but the estimated latent positions only lie near it.  Section~\ref{isomap} develops a new convergence analysis of Isomap in this setting.

Because Isomap approximates geodesic distances with Euclidean distances, the $1$-dimensional manifold that it learns is automatically parametrized by arc length.  Hence, we can approximate the sample Fr\'{e}chet mean in ${\mathcal H}$ with a conventional sample mean in the learnt manifold, and arc length distance with Euclidean distance.  The resulting test statistic is constructed as follows.
\begin{enumerate}

\item  To learn ${\mathcal H}$, apply Isomap to $p_0,\W \hat{X}_1,\ldots,\W \hat{X}_n$.  
\begin{enumerate}

\item Construct a localization graph.  While it is easier to develop theory for $\lambda$-neighborhood graphs, $K$-nearest-neighbor graphs are often preferred in practice.

\item Compute shortest path distances $\Delta = [ \delta_{ij} ]$ on the localization graph, thereby approximating geodesic distances on ${\mathcal H}$.  

\item Embed $\Delta$,  
obtaining $\hat{Z}_0,\hat{Z}_1,\ldots,\hat{Z}_n \in \Re$.  
To ensure that the Euclidean distances between these points approximate the shortest path distances (and therefore the geodesic distances on ${\mathcal H}$),
we prefer to embed $\Delta$ by minimizing the raw stress criterion.

\end{enumerate}

\item Set
$\bar{Z}_s = \frac{1}{s} \sum_{i=1}^{s} \hat{Z}_{i}$.

\item To approximate $T_1(\A)$, set
$\hat{{T}}_1 (\A) = \left| \bar{Z}_s - \hat{Z}_0 \right|$.
Critical regions of this test are of the form
$\{ \A : \hat{T}_1(\A) \geq c > 0 \}$.

\end{enumerate}

\subparagraph{Example 1 (continued)}
Set $s=5$ and $m=1000$.
Suppose that $p_1= \cdots = p_s = p^* = \psi(\tau^*)$
and that we wish to test $H_0: p^* = p_0 = \psi(0.3)$ at significance level $\alpha = 0.05$ using each of the three preceding tests with $\L=\I$.
For $\hat{T}_1$, we construct the localization graphs used in Isomap by connecting $x_i$ and $x_j$ if and only if $\| x_i-x_j \| \leq \lambda$, choosing $\lambda=1$.\footnote{This is a fairly large value of $\lambda$ in relation to the length of the Hardy-Weinberg submanifold, but it ensures that all of the localization graphs constructed in our numerical simulations are connected.  In practice, we would be inclined to use a smaller value, e.g., the smallest value for which the localization graph is connected.} To embed the shortest path distances in $\Re^1$, we use the \mbox{\tt R} package \mbox{\tt smacof} to initialize by CMDS and perform iterations of the Guttman transform with $u_{ij}=1$. 

We use Monte Carlo simulation to investigate the power of these tests at the alternative $p_a = \psi(0.35)$.   First, for $b=1,\ldots,1000$, we generate adjacency matrices $\A_0^b$ and corresponding test statistic values $T_k(\A_0^b)$, $T_1(\A_0^b)$, and $\hat{T}_1(\A_0^b)$ under the null RDPG probability model.  The corresponding (approximate) critical values, $C_k$, and $C_1$, $\hat{C}_1$ are the $0.95$ quantiles of the $1000$ observed values of the test statistics.

Next, for $b=1,\ldots,1000$, we generate adjacency matrices $\A^b$ and corresponding test statistic values  
$T_k(\A^b)$, $T_1(\A^b)$, and $\hat{T}_1(\A^b)$ under the alternative RDPG probability model.  
To estimate the power of each test, we count the fraction of times that we observe a test statistic value at least as great as its critical value, obtaining the following estimates:
\begin{eqnarray*}
\# \left\{ T_k \left( \A^b \right) \geq C_k \right\}/1000 & = & 0.633 \\
\# \left\{ T_1 \left( \A^b \right) \geq C_1 \right\}/1000 & = & 0.807  \\
\# \left\{ \hat{T}_1 \left( \A^b \right) \geq \hat{C}_1 \right\}/1000 & = & 0.960 
\end{eqnarray*}
The results are striking.  
First, they suggest that the restricted tests are indeed more powerful than the unrestricted test.  Second, the restricted test based on the learnt submanifold appears to be at least as powerful as the restricted test based on the known submanifold.  Such a conclusion would be of enormous consequence, as in practice the submanifold will be unknown.  Sections~\ref{isomap} and~\ref{power} provide theoretical justification for the efficacy of the restricted test based on the learnt submanifold. 
\hfill $\Box$

\section{Two-Sample Tests}
\label{test2}

In the context of random dot product graphs, $1$-sample problem are somewhat contrived.  Because latent positions are invariant under rotation, specification of the null hypothesis implies knowledge of a particular coordinate system that can be recovered by Procrustes analysis.  We have ignored this difficulty, assuming knowledge of the relevant rotation.  In practice, one is far more likely to encounter a $2$-sample problem, e.g., testing the null hypothesis that two Fr\'{e}chet means are identical.  For such problems, the hypotheses can be specified and the test statistics $T_k$ and $\hat{T}_1$ can be computed without reference to a particular coordinate system.

Suppose that $\A$ is an $n \times n$ adjacency matrix generated by an RDPG whose latent positions lie in ${\mathcal H}$.
Assume that the latent positions are of three types:
\begin{enumerate}

\item  A small number ($s_1$) of latent positions $p_1,\ldots,p_{s_1}$ with a fixed location $p_1^* = \psi(\tau_1^*)$, or sampled from a small neighborhood thereof.  Formally, let $h_1^*$ be a probability distribution on ${\mathcal H}$ with Fr\'{e}chet mean $p_1^*$ and assume that $p_1,\ldots,p_{s_1} \iid h_1^*$.

\item  A small number ($s_2$) of latent positions $p_{s_1+1},\ldots,p_{s_1+s_2}$ with a  
fixed location $p_2^* = \psi(\tau_2^*)$, or sampled from a small neighborhood thereof.  Formally, let $h_2^*$ be a probability distribution on ${\mathcal H}$ with Fr\'{e}chet mean $p_2^*$ and assume that $p_{s_1+1},\ldots,p_{s_1+s_2} \iid h_2^*$.
We hope to test the null hypothesis $H_0: p_1^* = p_2^*$ against the alternative hypothesis $H_1 : p_1^* \neq p_2^*$.  

\item  A large number ($m$) of auxiliary latent positions $p_{s_1+s_2+1},\ldots,p_{s_1+s_2+m}$ that provide information about the structure of ${\mathcal H}$.  Let $p_i = \psi(\tau_i)$.  We assume that $\tau_{s_1+s_2+1},\ldots,\tau_{s_1+s_2+m} \iid \mu$, where $\mu$ is a strictly positive probability density function on $[0,1]$.

\end{enumerate}
Collectively, $n=s_1+s_2+m$, the $n \times k$ latent position matrix is
\[
\X = \left[ \begin{array}{c|c|c|c|c|c|c|c|c} p_1 & \cdots & p_{s_1} & p_{s_1+1} & \cdots & p_{s_1+s_2} & p_{s_1+s_2+1} & \cdots & p_{s_1+s_2+m} \end{array} \right]^\top,
\]
and $\A \sim \mbox{RDPG}(\X)$.  Again $m \gg s_1,s_2$ and we propose tests that attempt to exploit the structure of ${\mathcal H}$.

As in Section~\ref{test1},
we observe $\A \sim \mbox{RDPG}(\X)$ and construct $\hat{X}_1,\ldots,\hat{X}_n \in \Re^r$,
then compute various test statistics that depend on $\hat{\X}$. 
Again we finesse the issue of how to determine $r$, assuming that $r=k$ is known.  In contrast to Section~\ref{test1}, however, the fact that the true latent positions are not identifiable does not pose a problem.  For $1$-sample problems, $\hat{\X}$ must be rotated to the representation in which the null hypothesis is specified.  For $2$-sample problems, the test statistic $T_1$ requires rotation to the representation in which the true submanifold is known; however, the test statistics $T_k$ and $\hat{T}_1$  are derived directly from $\hat{\X}$ and are invariant under rotation.

\subsection{Unrestricted Testing}
\label{unrestricted2}

Let
\begin{eqnarray*}
\bar{X}_1 = \frac{1}{s_1} \sum_{i=1}^{s_1} \hat{X}_i & \mbox{ and } &
\bar{X}_2 = \frac{1}{s_2} \sum_{i=s_1+1}^{s_1+s_2} \hat{X}_i
\end{eqnarray*}
denote the centroids of the estimated latent positions for the two small communities of interest.
Notice that rotation of $\hat{\X}$ is unnecessary.
For any symmetric positive definite $k \times k$ matrix $\L$,
define the unrestricted test statistic
\[
{T}_k^2(\A) = \left( \bar{X}_1 - \bar{X}_2 \right)^\top \L
\left( \bar{X}_1 - \bar{X}_2 \right).
\]
Critical regions of such tests are of the form
$\{ \A : {T}_k(\A) \geq c > 0 \}$.

If $\L=\I$, then
${T}_k(\A) = \| \bar{X}_1 - \bar{X}_2 \|_2$ is
the Euclidean distance in $\Re^k$ between the centroids.  
If $\L = {\mathbf \Sigma} \left( p_1^* \right)^{-1} =
{\mathbf \Sigma} \left( p_2^* \right)^{-1}$, then Theorem~\ref{thm:CLT} suggests that
${T}_k(\A)$ is approximately the Mahalanobis distance between the centroids.
Estimating this $\L$ by the pooled sample covariance matrix of $\hat{X}_1,\ldots,\hat{X}_{s_1}$ and $\hat{X}_{s_1+1},\ldots,\hat{X}_{s_1+s_2}$ leads to Hotelling's $2$-sample $T^2$ test.
Of course, if the assumption of equal covariance structures is not warranted, then one might prefer the tests of James \cite{James:1954} or Yao \cite{Yao:1965}.

\subsection{Testing Restricted to the True Submanifold}
\label{restricted2}

Suppose that $p_1,\ldots,p_n \in {\mathcal H} = \psi([0,1])$, where $\psi : [0,1] \rightarrow \Re^k$ is known. 
For a fixed symmetric positive definite $k \times k$ matrix $\L$,
we estimate $p_i=\psi(\tau_i)$ by 
$\hat{p}_i=\psi(\hat{\tau}_i)$, where
$\hat{\tau}_i$, minimizes the objective function
\[
\mbox{MDE}(\tau) = \left[ \psi(\tau) - \hat{\W} X_i \right]^\top
\L^{-1}
\left[ \psi(\tau) - \hat{\W} X_i \right].
\]
Notice that computing $\mbox{MDE}(\tau)$ requires knowledge of the orthogonal matrix $\hat{\W}$ that rotates $\hat{\X}$ to the representation in which ${\mathcal H} = \psi([0,1])$ is specified.

Assuming that $\| \dot{\psi} (\tau) \|$ is bounded,
the restricted test statistic $T_1(\A)$ is the geodesic distance $d_M$ between the sample Fr\'{e}chet means of 
\begin{eqnarray*}
\psi(\hat{\tau}_1),\ldots,\psi(\hat{\tau}_{s_1}) & \mbox{ and } & \psi(\hat{\tau}_{s_1+1}),\ldots,\psi(\hat{\tau}_{s_1+s_2}).
\end{eqnarray*} 
To compute $T_1(\A)$, first compute
\[
\hat{t}_i = \int_0^{\hat{\tau}_i} 
\left\| \dot{\psi}(\tau) \right\| \, d\tau, \quad
\bar{t}_1 = \frac{1}{s_1} \sum_{i=1}^{s_1} \hat{t}_i,
\quad \mbox{and} \enspace
\bar{t}_2 = \frac{1}{s_2} \sum_{i=s_1+1}^{s_1+s_2} \hat{t}_i. 
\]
Then
\begin{eqnarray*}
{T}_1 ( \A ) & = & d_M \left(
\mu_{\mbox{\rm \tiny Fr}} \left( 
\psi(\hat{\tau}_1),\ldots,\psi(\hat{\tau}_{s_1}) \right) , 
\mu_{\mbox{\rm \tiny Fr}} \left( 
\psi(\hat{\tau}_{s_1+1}),\ldots,\psi(\hat{\tau}_{s_1+s_2}) \right)
\right) \\ & = &
\left| \bar{t}_1-\bar{t}_2 \right| =
\left| \frac{1}{s_1s_2} 
 \sum_{i=1}^{s_1} \sum_{j=s_1+1}^{s_1+s_2} 
 \int_{\hat{\tau}_j}^{\hat{\tau}_i} 
 \left\| \dot{\psi}(\tau) \right\| \, d\tau \right|.
\end{eqnarray*}
Critical regions of these tests are of the form
$\{ \A : {T}_1(\A) \geq c > 0 \}$.

\subsection{Testing Restricted to a Learnt Submanifold}
\label{restricted2.learnt}

If ${\mathcal H}$ is unknown, 
then the restricted test statistic $\hat{T}_1$ is constructed as follows.
\begin{enumerate}

\item  Apply Isomap to $\hat{X}_1,\ldots, \hat{X}_n$.
Notice that rotation of $\hat{\X}$ is unnecessary. 
\begin{enumerate}

\item Construct a localization graph.

\item Compute shortest path distances $\Delta = [ \delta_{ij} ]$ on the localization graph.  

\item Embed $\Delta$ by minimizing the raw stress criterion,  
obtaining $\hat{Z}_1,\ldots,\hat{Z}_n \in \Re$.  

\end{enumerate}

\item Set
\begin{eqnarray*}
\bar{Z}_1 = \frac{1}{s_1} \sum_{i=1}^{s_1} \hat{Z}_{i} & 
\mbox{ and } &
\bar{Z}_2 = \frac{1}{s_2} \sum_{i=s_1+1}^{s_1+s_2} \hat{Z}_{i}.
\end{eqnarray*}

\item Set
$\hat{{T}}_1 (\A) = \left| \bar{Z}_1 - \bar{Z}_2 \right|$.
Critical regions of this test are of the form
$\{ \A : \hat{T}_1(\A) \geq c > 0 \}$.

\end{enumerate}

\section{Isomap Convergence Analysis}
\label{isomap}

The behavior of the restricted test based on the learnt submanifold depends on the behavior of Isomap as the number of vertices in the approximating graph increases.
In what follows, we appropriate several key elements of the analysis of Isomap described in \cite{Bernstein&etal:2000}.
Notice, however, that the $\W_n \hat{X}_{ni}$ from which the approximating graphs are constructed do not lie on the submanifold
${\mathcal H}$.  The authors of \cite{Bernstein&etal:2000} left ``a formal analysis of Isomap with noisy data to future work.''  So far as we are aware, what follows is the first attempt at such an analysis.  

\subsection{Manifold Structure}

Let ${\mathcal M} \subset \Re^k$ denote a $1$-dimensional compact Riemannian manifold. 
Suppose that ${\mathcal M} = \gamma([0,L])$ with 
$\left\| \dot{\gamma}(t) \right\| =1$, 
i.e., $\gamma$ is parametrized by arc length.
Let $d_M$ denote arc length (geodesic) distance on ${\mathcal M}$, i.e.,
\begin{equation}
d_M \left( \gamma \left( t_1 \right), \gamma \left( t_2 \right) \right) 
= \left| \int_{t_1}^{t_2} \| \dot{\gamma}(t) \| \, dt  \right|
= \left| t_2-t_1 \right|.
\label{eq:arclength}
\end{equation}

Following \cite{Bernstein&etal:2000},
suppose that there exists $r>0$ such that $\left\| \ddot{\gamma} (t) \right\| \leq 1/r$.
Let $r_0$ denote the {\em minimum radius of curvature}\/ of ${\mathcal M}$, 
the largest $r$ for which this inequality holds.
Let $s_0$ denote the
{\em minimum branch separation}\/ of ${\mathcal M}$, i.e.,
the largest $s$ for which $\| x-y \| < s$ entails $d_M(x,y) \leq \pi r_0$ for every $x,y \in {\mathcal M}$.  The quantity $s_0$ has also been called the {\em proximity to self-intersection}.

For $\sigma \geq 0$, let
\[
{\mathcal M}_\sigma = \left\{ x \in \Re^k \; : \;
\min_{z \in {\mathcal M}} \| x-z \| \leq \sigma \right\}.
\]
The ${\mathcal M}_\sigma$ are nested, i.e., $\sigma_1 < \sigma_2$ entails ${\mathcal M}_{\sigma_1} \subset {\mathcal M}_{\sigma_2}$,
with ${\mathcal M}_0 = {\mathcal M}$.
If $\sigma < s_0/3$, then ${\mathcal M}_\sigma$ does not self-intersect and is itself a $k$-dimensional compact Riemannian manifold with minimum branch separation at least $s_0/3$.
Let $d_\sigma$ denote geodesic distance on ${\mathcal M}_\sigma$.
If $x,y \in {\mathcal M} \subset {\mathcal M}_\sigma$, then
$d_\sigma(x,y) \leq d_M(x,y)$ and 
\[
\lim_{\sigma \rightarrow 0} d_\sigma(x,y) = d_M(x,y).
\]

\subsection{Graph Structure}

Suppose that $\delta,\epsilon,\sigma >0$ satisfy 
$2 \delta \leq \epsilon$ and 
$\sigma = \delta + \epsilon/2 < s_0/3$.
Suppose that $x_1,\ldots,x_m \in {\mathcal M}$, with $x_i = \gamma(t_i)$ for $t_1 \leq \cdots \leq t_m$.  Suppose that every $x \in {\mathcal M}$ lies within arc length $\delta$ of some $x_i$.  Let $V = \{ \hat{x}_1,\ldots,\hat{x}_m \} \subset \Re^k$ be such that each $\hat{x}_i$ lies within $\delta$ of $x_i$, so that $V \subset {\mathcal M}_\delta$.  
We emphasize that we do {\em not}\/ assume that $V$ lies in ${\mathcal M}$.
Let ${\mathcal G}$ denote the $(\epsilon+2\delta)$-neighborhood graph with vertex set $V$, i.e., vertices $\hat{x}_i$ and $\hat{x}_j$ are connected by an edge if and only if $\| \hat{x}_i-\hat{x}_j \| \leq \epsilon+2\delta$.
The line segment connecting $\hat{x}_i$ and $\hat{x}_j$ may not lie in ${\mathcal M}_\delta$, but it cannot lie farther away than $(\epsilon+2\delta)/2$.  (Sharper bounds are possible, but not needed for our analysis.)  Hence, ${\mathcal G} \subset {\mathcal M}_\sigma$.  
To see that ${\mathcal G}$ is connected, note that
\begin{eqnarray} \nonumber
\left\| \hat{x}_{i-1}-\hat{x}_{i} \right\| & \leq &
\left\| \hat{x}_{i-1}-x_{i-1} \right\| +
\left\| x_{i-1}-x_{i} \right\| +
\left\| x_{i}-\hat{x}_{i} \right\| \\ & \leq &
\delta + d_M \left( x_{i-1},x_{i} \right) + \delta \leq 
4 \delta \leq \epsilon+2\delta
\label{eq:connect}
\end{eqnarray}
for $i=2,\ldots,m$.

Let $d_G$ denote shortest path distance on ${\mathcal G}$.
If $\hat{x}_a,\hat{x}_b \in V$, then
$d_G \left( \hat{x}_a,\hat{x}_b \right) \geq
d_\sigma \left( \hat{x}_a,\hat{x}_b \right)$.
Furthermore,
\[
d_\sigma \left( x_a,x_b \right) \leq
d_\sigma \left( x_a,\hat{x}_a \right) +
d_\sigma \left( \hat{x}_a,\hat{x}_b \right) +
d_\sigma \left( \hat{x}_b,x_b \right) =
\delta + d_\sigma \left( \hat{x}_a,\hat{x}_b \right) + \delta,
\]
so that
\begin{equation}
d_G \left( \hat{x}_a,\hat{x}_b \right) \geq
d_\sigma \left( \hat{x}_a,\hat{x}_b \right) \geq
d_\sigma \left( x_a,x_b \right) - 2\delta
\label{eq:lower}
\end{equation}
provides a lower bound on shortest path distance.

To obtain an upper bound, first let $\ell = d_M(x_a,x_b)$ and suppose that $j=2\ell/\epsilon$ is an integer.  
Label $x_a$ and $x_b$ so that $x_a = \gamma(t_0)$ and $x_b = \gamma(t_j)$ with $t_0 < t_j$.
For $i=1,\ldots,j$, set
$t_i=t_0+i\epsilon/2$ and $I_i = [t_{i-1},t_i]$.
Choose $x_i \in \{ x_1,\ldots,x_m \} \cap I_i$ and notice that
\[
\ell = d_M \left( x_a,x_b \right) = 
\sum_{i=1}^j d_M \left( x_{i-1},x_i \right).
\]
It follows from (\ref{eq:connect})
that the vertices $\hat{x}_{i-1}$ and $\hat{x}_i$ are connected by an edge in ${\mathcal G}$, hence that
\[
\hat{x}_0 \leftrightarrow \hat{x}_1 \leftrightarrow \cdots
\leftrightarrow \hat{x}_{j-1} \leftrightarrow \hat{x}_j.
\]
is a path from $\hat{x}_a$ to $\hat{x}_b$.
As a result,
\begin{eqnarray} \nonumber
d_G \left( \hat{x}_a,\hat{x}_b \right) & \leq &
\sum_{i=1}^j \left\| \hat{x}_{i-1} - \hat{x}_i \right\| \leq
\sum_{i=1}^j \left[ d_M \left( x_{i-1},x_k \right) + 2\delta \right] \\ \nonumber & \leq &
d_M \left( x_a,x_b \right) + 2j\delta = 
d_M \left( x_a,x_b \right) + 2(2\ell/\epsilon)\delta \\  & = &
\left( 1 + \frac{4\delta}{\epsilon} \right) d_M \left( x_a,x_b \right) .
\label{eq:upper}
\end{eqnarray}

\subsection{Probability Structure}

To obtain a sufficiently dense sample $x_1,\ldots,x_m \in {\mathcal M} = \gamma([0,L])$, we assume that $x_i = \gamma(t_i)$, where $t_1,\ldots,t_m \iid \nu$.  The following lemma is analogous to the Sampling Lemma in \cite{Bernstein&etal:2000}.
\begin{lemma}
Suppose that the probability density function $\nu : [0,L] \rightarrow \Re$ has minimum value $\nu_{min}>0$ and that 
$t_1,\ldots,t_m \iid \nu$.  
Let $\ell$ be a natural number such that $\delta=L/\ell \in (0,1/\nu_{min})$.
Let $E_m$ denote the event that every $x \in \gamma([0,L])$ lies within arc length $\delta$ of some $x_j = \gamma(t_j)$.  Then
$\lim_{m \rightarrow \infty} P(E_m)=1$.
\label{lm:sampling}
\end{lemma}

\subparagraph{Proof}
Partition $[0,L]$ into intervals $I_1,\ldots,I_\ell$, each of length $\delta$, and set $B_i = \gamma(I_i)$. 
If each $B_i$ contains at least one $x_j$, then $E_m$ obtains.
In fact,
\[
\delta = 
\int_{I_i} 1 \, dt
\leq \frac{1}{\nu_{min}} 
\int_{I_i} \nu(\tau) \, dt
\]
and the probability that each $B_i$ contains at least one $x_j = \gamma(t_j)$ is
\begin{eqnarray*}
P \left( \mbox{every $I_i$ contains a $t_j$} \right)
 & = & 1- P \left( \mbox{some $I_i$ contains no $t_j$} \right) \\ 
  & \geq & 1- \sum_{i=1}^\ell P \left( \mbox{$I_i$ contains no $t_j$} \right) \\
  & = & 1- \sum_{i=1}^\ell \prod_{j=1}^m P \left( t_j \not\in I_i \right) \\
  & = & 1- \sum_{i=1}^\ell \prod_{j=1}^m \left(
  1 - \int_{I_i} \nu(t) \, dt \right) \\
  & \geq & 1- \sum_{i=1}^\ell \prod_{j=1}^m \left(
  1 - \nu_{min} \delta \right) \\
  & = & 1-\ell \left(
  1 - \nu_{min} \delta \right)^m,
\end{eqnarray*}
which tends to $1$ as $m \rightarrow \infty$.
\hfill $\Box$

\subsection{Convergence of Shortest Path Distances}
\label{SPDconverge}
Main Theorem~B in \cite{Bernstein&etal:2000} requires data that lie on the manifold to be learned, i.e., $x_1,\ldots,x_m \in {\mathcal M}$. Combining the preceding,
we obtain an analogous result with data that approach the manifold asymptotically, i.e., $\hat{x}_1,\ldots,\hat{x}_m \in {\mathcal M}_\delta$.

\begin{theorem}
Suppose that $\gamma : [0,L] \rightarrow \Re^k$ is such that
$\left\| \dot{\gamma} (t) \right\| =1$ and $\left\| \ddot{\gamma} (t) \right\| \leq 1/r_0 < \infty$.
Let $d_M$ denote arc length distance on the $1$-dimensional compact Riemannian manifold ${\mathcal M}=\gamma([0,L])$.  

Suppose that the probability density function $\nu : [0,L] \rightarrow \Re$ has minimum value $\nu_{min}>0$, and
that $t_1,\ldots,t_m \iid \nu$.
Let $x_i = \gamma(t_i)$, and suppose that $\| \hat{x}_i - x_i \| < \delta_K$.

Let $d_{m,\lambda}$ denote shortest path distance on ${\mathcal G}_{m,\lambda}$, the $\lambda$-neighborhood graph constructed from $\hat{x}_1,\ldots,\hat{x}_m$.  If $\delta_K \rightarrow 0$, then there exist corresponding sequences of neighborhood sizes $\lambda_K \rightarrow 0$ and sample sizes $m_K \rightarrow \infty$ such that
$d_{m_K,\lambda_K} \left( 
\hat{x}_a,\hat{x}_b \right)$ converges in probability to 
$d_M \left( x_a,x_b \right)$
for every $(x_a,x_b)$. 
\label{thm:isomapB}
\end{theorem}

\subparagraph{Proof}
Choose $\epsilon_K$ so that $\epsilon_K \rightarrow 0$ and $\delta_K/\epsilon_K \rightarrow 0$. 
For $K$ sufficiently large,
$2 \delta_K \leq \epsilon_K$ and 
$\sigma_K = \delta_K + \epsilon_K/2 < s_0/3$,
where $s_0$ is the minimum branch separation of ${\mathcal M}$.
As $K \rightarrow \infty$, both $\lambda_K = \epsilon_K+2\delta_K \rightarrow 0$ and $\sigma_K \rightarrow 0$.

Suppose that $\pi_K \rightarrow 0$ is a decreasing sequence of error probabilities.
Let $E_K$ denote the event that every $x \in {\mathcal M}$ lies within arc length $\delta_K$ of some $x_j \in \{ x_1,\ldots,x_{m_K} \}$, and
apply Lemma~\ref{lm:sampling} to choose $m_K$ so that $P(E_K) \geq 1-\pi_K$.

If $E_K$ obtains, then it follows from (\ref{eq:lower}) that
\[
\lim_{K \rightarrow \infty} 
d_{m_K,\lambda_K} \left( \hat{x}_a,\hat{x}_b \right) \geq
\lim_{K \rightarrow \infty}
d_{\sigma_K} \left( x_a,x_b \right) - 2\delta_K =
d_M \left( x_a,x_b \right).
\]
Furthermore, it follows from (\ref{eq:upper}) that
\[
\lim_{K \rightarrow \infty} 
d_{m_K,\lambda_K} \left( \hat{x}_a,\hat{x}_b \right) \leq
\lim_{K \rightarrow \infty}
\left( 1 + \frac{4\delta_K}{\epsilon_K} \right) d_M \left( x_a,x_b \right) = d_M \left( x_a,x_b \right).
\]
Because $1-\pi_K \rightarrow 1$ as $K \rightarrow \infty$,
we conclude that 
$d_{m_K,\lambda_K} \left( 
\hat{x}_a,\hat{x}_b \right)$ converges in probability to 
$d_M \left( x_a,x_b \right)$.
\hfill $\Box$

\subsection{Convergence of Euclidean Distances}
\label{EDconverge}
From (\ref{eq:arclength}), 
$\gamma^{-1}({\mathcal M}) = [0,L] \subset \Re$ is a $1$-dimensional embedding of ${\mathcal M}=\gamma([0,L])$ with the property that Euclidean distance in $[0,L]$ equals geodesic distance in ${\mathcal M}$.  Thus, if $t_1,\ldots,t_m \in [0,L]$, then
\[
\frac{1}{2} \sum_{i,j=1}^m  \left[ \left| t_i-t_j \right| -  d_M \left( \gamma \left( t_i \right), \gamma \left( t_j \right) \right) \right]^2 = 0.
\]
Under the assumptions of Theorem~\ref{thm:isomapB}, suppose that $\hat{z}_1,\ldots,\hat{z}_\ell \in \Re$ minimize the unweighted raw stress criterion,
\[
\sigma_\ell \left( z_1,\ldots,z_\ell \right) =
\frac{1}{2} \sum_{i,j=1}^\ell  \left[ \left| z_i-z_j \right| -  d_{m,\lambda} \left( \hat{x}_i,\hat{x}_j \right) \right]^2,
\]
where $\ell \leq m$.
Because each $d_{m,\lambda} \left( \hat{x}_i,\hat{x}_j \right)$ converges in probability to $d_M \left( \gamma \left( t_i \right), \gamma \left( t_j \right) \right) = \left| t_i-t_j \right|$,
it is plausible that the $\left| \hat{z}_i-\hat{z}_j \right|$ converge to the $\left| t_i-t_j \right|$ in a suitable sense.

There are (at least) three different embedding schemes that warrant investigation.  Let 
$\Delta(\ell,m) = [ d_{m,\lambda} (\hat{x}_i,\hat{x}_j) ] = [ \delta_{ij} ]$ for $i,j=1,\ldots,\ell$.
We might
\begin{enumerate}

\item Fix $\ell$, i.e., embed $\Delta(\ell,m)$ and let $m \rightarrow \infty$.  
This scheme will suffice for the asymptotic power analyses in Section~\ref{power}.

\item Allow $\ell \rightarrow \infty$, but not require $\ell = m$.

\item Require $\ell = m$, i.e., embed $\Delta(m,m)$ and let $m \rightarrow \infty$.  This is the most natural way to embed when $m$ is fixed, as is typical in applications.  However, we will defer asymptotic analysis of this scheme to future work, as it requires a stronger convergence result than Theorem~\ref{thm:isomapB}.

\end{enumerate}

We begin by characterizing minimizers of the raw stress criterion when $d=1$.
\begin{lemma}
Let $\Delta = [ \delta_{ij} ]$ be an $\ell \times \ell$ dissimilarity matrix with strictly positive off-diagonal entries.  Let 
$\hat{z}_1,\ldots,\hat{z}_\ell \in \Re$ with $\sum_{i=1}^\ell \hat{z}_i = 0$ be a minimizer of $\sigma_\ell ( z_1,\ldots,z_\ell )$.  Then
\[
\hat{z}_k =  \frac{1}{\ell} \left[
\sum_{i \in I_k} \delta_{ik} - \sum_{i \in J_k} \delta_{ik} \right],
\]
where $I_k = \{ i : \hat{z}_i < \hat{z}_k \}$ and
$J_k = \{ i : \hat{z}_i > \hat{z}_k \}$.
\label{lm:mds1}
\end{lemma}

\subparagraph{Proof}
It was established in \cite{deleeuw:1984} that $\sigma$ is differentiable at $(\hat{z}_1,\ldots,\hat{z}_\ell)$, and that the $\hat{z}_i$ are necessarily distinct.
Choose the indexing for which $\hat{z}_1 < \cdots < \hat{z}_\ell$,
in which case $I_k = \{ 1,\ldots,k-1 \}$ and $J_k = \{ k+1,\ldots,\ell \}$, then write
\[
\sigma_\ell \left( z_1,\ldots,z_\ell \right) =
\sum_{i=2}^\ell \sum_{j=1}^{i-1} \left( z_i-z_j-\delta_{ij} \right)^2,
\]
compute partial derivatives
\begin{eqnarray*}
\frac{\partial}{\partial z_k} \sigma \left( z_1,\ldots,z_\ell \right)
 & = &
\frac{\partial}{\partial z_k} \left[
\sum_{j=1}^{k-1} \left( z_k-z_j-\delta_{kj} \right)^2 +
\sum_{i=k+1}^\ell \left( z_i-z_k-\delta_{ik} \right)^2 \right] \\
 & = &
2 \sum_{j=1}^{k-1} \left( z_k-z_j-\delta_{kj} \right) -
2 \sum_{i=k+1}^\ell \left( z_i-z_k-\delta_{ik} \right) \\
 & = &
2(k-1)z_k - 2 \sum_{j=1}^{k-1} z_j - 2 \sum_{j=1}^{k-1} \delta_{kj} + \\ & & 2(\ell-k)z_k
-2 \sum_{i=k+1}^\ell z_i + 2 \sum_{i=k+1}^\ell \delta_{ik} \\
 & = &
2(\ell-1)z_k -2 \left[ \sum_{j=1}^{k-1} z_j + \sum_{j=k+1}^\ell z_j \right] -2 \left[
\sum_{i=1}^{k-1} \delta_{ik} - \sum_{i=k+1}^\ell \delta_{ik} \right] \\
 & = &
2\ell z_k -2 \left[
\sum_{i=1}^{k-1} \delta_{ik} - \sum_{i=k+1}^\ell \delta_{ik} \right],
\end{eqnarray*}
and conclude that
\[
\hat{z}_k = \frac{1}{\ell} \left[
\sum_{i=1}^{k-1} \delta_{ik} - \sum_{i=k+1}^\ell \delta_{ik} \right].
\]
\hfill $\Box$

\begin{theorem}
Fix $\ell \geq 2$ and embed $\Delta(\ell,m) = [\delta_{ij}]$ by minimizing $\sigma_\ell$, obtaining a centered configuration $\hat{z}_1,\ldots,\hat{z}_\ell \in \Re$.  Fix $a,b \in \{ 1,2,\ldots \}$.  
If $a > \ell$, then set
\begin{equation}
\hat{z}_a = \frac{1}{\ell(\ell-1)} \sum_{i<j}
\frac{\delta_{ja}^2-\hat{z}_j^2-\delta_{ia}^2+\hat{z}_i^2}{
\hat{z}_i - \hat{z}_j},
\label{eq:out}
\end{equation}
and likewise for $b$.
Then, as $K \rightarrow \infty$ under the conditions of Theorem~\ref{thm:isomapB}, $\left| \hat{z}_a-\hat{z}_b \right|$
converges in probability to $d_M \left( x_a,x_b \right)$.
\label{thm:converge.ED}
\end{theorem}

\subparagraph{Proof}
First suppose that $a,b \leq \ell$.
Using Lemma~\ref{lm:mds1} and holding $\ell$ fixed as $m \rightarrow \infty$, we obtain 
\begin{eqnarray*}
\lefteqn{\lim_{m \rightarrow \infty}
\left| \hat{z}_a \left( \Delta(\ell,m) \right) -
\hat{z}_b \left( \Delta(\ell,m) \right) \right|} & & \\ & = &
\lim_{m \rightarrow \infty}
\frac{1}{\ell} \left|
\sum_{i \in I_a} \delta_{ia} - \sum_{i \in J_a} \delta_{ia} -
\sum_{i \in I_b} \delta_{ib} + \sum_{i \in J_b} \delta_{ib}
\right| \\ & = &
\frac{1}{\ell} \left|
\sum_{i \in I_a} \left( t_a-t_i \right) - \sum_{i \in J_a} \left( t_i-t_a \right) -
\sum_{i \in I_b} \left( t_b-t_i \right) + \sum_{i \in J_b} \left( t_i-t_a \right)
\right| \\ & = &
\left| t_a-t_b \right|.
\end{eqnarray*}

Now suppose that $a,b > \ell$.  Then, using the previous result,
\begin{eqnarray*}
\lefteqn{\lim_{m \rightarrow \infty}
\left| \hat{z}_a \left( \Delta(\ell,m) \right) -
\hat{z}_b \left( \Delta(\ell,m) \right) \right|} & & \\ & = &
\lim_{m \rightarrow \infty} \left|
\frac{1}{\ell(\ell-1)} \sum_{i<j} \frac{
\delta_{ja}^2-\hat{z}_j^2-\delta_{ia}^2+\hat{z}_i^2-
\delta_{jb}^2+\hat{z}_j^2+\delta_{ib}^2-\hat{z}_i^2}{
\hat{z}_i - \hat{z}_j} \right| \\ & = &
\lim_{m \rightarrow \infty} \left|
\frac{1}{\ell(\ell-1)} \sum_{i<j} \frac{
\delta_{ja}^2-\delta_{ia}^2-
\delta_{jb}^2+\delta_{ib}^2}{
\hat{z}_i - \hat{z}_j} \right| \\ & = &
\left|
\frac{1}{\ell(\ell-1)} \sum_{i<j} \frac{
(t_j-t_a)^2-(t_i-t_a)^2-(t_j-t_b)^2+(t_i-t_b)^2}{
t_i-t_j} \right| \\ & = &
\left|
\frac{1}{\ell(\ell-1)} \sum_{i<j} 2 \left( t_a-t_b \right) \right| \\ & = &
\left| t_a-t_b \right|.
\end{eqnarray*}
The remaining cases are analogous. \hfill $\Box$

\bigskip

Theorem~\ref{thm:converge.ED} relies on an embedding strategy sometimes known as the {\em method of standards}.  To embed $m$ objects, one first embeds $\ell < m$ objects, then uses them as standards (or landmarks) for {\em out-of-sample}\/ embedding of the remaining $m-\ell$ objects.  See \cite{kruskal&hart:1966} for an early example.  More recently, {\em Landmark MDS}\/ \cite{desilva&tenenbaum:2003,desilva&tenenbaum:2004}
was proposed for the purpose of reducing the computational expense of Isomap's embedding step.  Our implementation of the method of standards, which embeds the standards by minimizing the raw stress criterion and performs out-of-sample embedding by (\ref{eq:out}), differs slightly from Landmark MDS but is entirely in the same spirit.

It is not difficult to weaken Theorem~\ref{thm:converge.ED}, allowing the number of standards to increase.  Let $\sigma_{\ell,m}$ denote the global minimum of $\sigma_\ell$ when embedding $\Delta(\ell,m)$.  We then have the following double sequence of real numbers:
\[
\begin{array}{cccccccc}
       & m=1 & m=2 & m=3 & m=4 & m=5 & \cdots & \\
\ell=1 & \sigma_{1,1} & \sigma_{1,2} & \sigma_{1,3} & \sigma_{1,4} & \sigma_{1,5} & \rightarrow & b_1=0 \\
\ell=2 &  & \sigma_{2,2} & \sigma_{2,3} & \sigma_{2,4} & \sigma_{2,5} & \rightarrow & b_2=0 \\
\ell=3 &  &  & \sigma_{3,3} & \sigma_{3,4} & \sigma_{3,5} & \rightarrow & b_3=0 \\
\ell=4 &  &  &  & \sigma_{4,4} & \sigma_{4,5} & \rightarrow & b_4=0 \\
\ell=5 &  &  &  &  & \sigma_{5,5} & \rightarrow & b_5=0 \\
\vdots &  &  &  &  &  &  & \downarrow \\
 &  &  &  &  &  &  & c=0
\end{array}
\]
A simple argument \cite[Proposition 1]{Sekhon:2021} then establishes that there is a strictly increasing subsequence $m_\ell$ such that $\sigma_{\ell,m_\ell} \rightarrow 0$ as $\ell \rightarrow \infty$.  Expressed in terms of $m=m_K$ in Theorem~\ref{thm:converge.ED}, we can allow $\ell=\ell_K$ to increase, but perhaps not so rapidly as $m_K$.

To establish that $\sigma_{m,m} \rightarrow 0$, we must either show that the row sequences $\sigma_{\ell,m}$ converge uniformly, or find a different proof technique.  We leave these investigations for future research.

\section{Power Comparison of Restricted Tests}
\label{power}

We have proposed restricted test statistics in two cases:
the case that the true submanifold is known 
($T_1$ in Sections \ref{restricted} and \ref{restricted2})
and the case that the true submanifold is unknown and must be learned
($\hat{T}_1$ in Sections \ref{restricted.learnt} and \ref{restricted2.learnt}).  
Whereas the latter case typically obtains in practice,
the former case provides an idealized benchmark.  
We would like to know that, 
if the true submanifold can be learned reasonably well, 
then the performance of $\hat{T}_1$ will approximate the performance of $T_1$.  Toward that end,
we investigate the power of the $1$- and $2$-sample $\hat{T}_1$ tests as the number of auxiliary points used to learn the submanifold 
${\mathcal H}$ increases.

Notice that we are not conducting an asymptotic analysis in the traditional sense.  While $m \rightarrow \infty$, the sample sizes drawn from the distribution(s) about which we are drawing inferences ($s$ in Sections \ref{restricted} and \ref{restricted.learnt}, $s_1$ and $s_2$ in Sections \ref{restricted2} and \ref{restricted2.learnt}) are fixed.

For $1$-sample problems,
let $p^*=\psi(\tau^*)$ denote the Fr\'{e}chet mean of the probability distribution $h^*$ on the $1$-dimensional submanifold ${\mathcal H} = \psi([0,1])$.
For $s$ fixed, we wish to test $H_0:p^*=p_0$ at significance level $\alpha$.
For $2$-sample problems,
let $p_1^*=\psi(\tau_1^*)$ and $p_2^*=\psi(\tau_2^*)$ denote the Fr\'{e}chet means of the probability distributions $h_1^*$ and $h_2^*$ on the $1$-dimensional submanifold ${\mathcal H} = \psi([0,1])$.
For $s_1$ and $s_2$ fixed, we wish to test $H_0:p_1^*=p_2^*$ at significance level $\alpha$.
For either case,
let $\pi_1(\cdot;m)$ and $\hat{\pi}_1(\cdot;m)$ 
denote the power functions of level $\alpha$ tests of $H_0$ based on
the test statistics $T_1$ and $\hat{T}_1$ respectively with $m$ auxiliary latent positions.
We study the behavior of these power functions as $m \rightarrow \infty$.

\subsection{Restricted to the True Submanifold}
The case of $T_1$ is straightforward.
From Theorem~\ref{thm:2inf},
\[
\max_{i=1,\ldots,s} \left\| \W_n \hat{X}_{ni} - p_i \right\| 
\stackrel{P}{\rightarrow} 0
\]
as $m \rightarrow \infty$.
Recalling that minimum distance estimation is consistent
under standard regularity conditions \cite{basu&etal:2011},
we assume that $\hat{\tau}_i \stackrel{P}{\rightarrow} {\tau}_i$ as $\W_n \hat{X}_{ni} \stackrel{P}{\rightarrow} p_i =\psi(\tau_i)$.  

Define the arc length function $R : [0,1] \rightarrow \Re$ by
\[
R(b) = d_{\mathcal H}(\psi(0),\psi(b)) =
\int_0^b \left\| \dot{\psi}(\tau) \right\| \, d\tau
\]
and let $m \rightarrow \infty$.
Because $R$ is continuous, $R(\hat{\tau}_i) \stackrel{P}{\rightarrow} R({\tau}_i)$.
For $1$-sample problems, 
\[
T_1 \left( \A \right) =
\left| \frac{1}{s} \sum_{i=1}^s R \left( \hat{\tau}_i \right) - 
R \left( \tau_0 \right) \right|
\stackrel{P}{\rightarrow}
\left| \frac{1}{s} \sum_{i=1}^s R \left( \tau_i \right) -
R \left( \tau_0 \right) \right|
= \left|  \mu_{\mbox{\rm \tiny Fr}} 
\left( p_1,\ldots,p_s \right)-p_0 \right| ,
\]
a function of the random variables $p_1,\ldots,p_s \iid h^*$.
For $2$-sample problems, 
\begin{eqnarray*}
T_1 \left( \A \right) & = &
\left| \frac{1}{s_1s_2} 
 \sum_{i=1}^{s_1} \sum_{j=s_1+1}^{s_1+s_2} 
 R \left( \hat{\tau}_i \right) - R \left( \hat{\tau}_j \right)
 \right| \\
 & \stackrel{P}{\rightarrow} &
 \left| \frac{1}{s_1s_2} 
 \sum_{i=1}^{s_1} \sum_{j=s_1+1}^{s_1+s_2} 
 R \left( \tau_i \right) - R \left( \tau_j \right)
 \right| \\
  & = &
  \left|  \mu_{\mbox{\rm \tiny Fr}} 
\left( p_1,\ldots,p_{s_1} \right) - \mu_{\mbox{\rm \tiny Fr}}
\left( p_{s_1+1},\ldots,p_{s_1+s_2} \right) \right|,
\end{eqnarray*}
a function of the random variables $p_1,\ldots,p_{s_1} \iid h_1^*$ and $p_{s_1+1},\ldots,p_{s_1+s_2} \iid h_2^*$.

Now let $C_1(m)$ denote the $1-\alpha$ quantile of $T_1(\A)$ and let $C_1 = \lim_{m \rightarrow \infty} C_1(m)$.  Then
\[
\lim_{m \rightarrow \infty} \pi_1 \left( \tau^*;m \right) = 
\lim_{m \rightarrow \infty} P \left( T_1 \left( \A \right) \geq C_1(m) \right) =
P \left( \left| \mu_{\mbox{\rm \tiny Fr}} 
\left( p_1,\ldots,p_s \right) -p_0 \right|
 \geq C_1 \right),
\]
for $1$-sample problems, and
\begin{eqnarray*}
\lim_{m \rightarrow \infty} \pi_1 \left( \tau_1^*,\tau_2^*;m \right)
 & = &
\lim_{m \rightarrow \infty} P \left( T_1 \left( \A \right) \geq C_1(m) \right) \\
 & = &
P \left( \left| \mu_{\mbox{\rm \tiny Fr}} 
\left( p_1,\ldots,p_{s_1} \right) -
\mu_{\mbox{\rm \tiny Fr}} 
\left( p_{s_1+1},\ldots,p_{s_1+s_2} \right) 
\right| \right)
\end{eqnarray*}
for $2$-sample problems.

\subsection{Restricted to a Learnt Submanifold}
Finally we demonstrate that, as the number of auxiliary latent positions increases, the power of the restricted test on the learnt manifold ($\hat{T}_1$) tends to the power of the restricted test on the true manifold ($T_1$).
Our argument relies on the analysis of Isomap in Section~\ref{isomap}.
To apply these results to ${\mathcal H} = \psi([0,1])$
with bounded $\| \dot{\psi}(\tau) \|$, we reparametrize $\psi$ so that ${\mathcal H} = \gamma([0,L])$ with $\| \dot{\gamma}(t) \|=1$.  If $\mu$ is a strictly positive probability density function on $[0,1]$, then reparametrization induces a strictly positive probability density function $\nu$ on $[0,L]$.

\begin{theorem}
Suppose that ${\mathcal H} = \psi([0,1])$ is a smooth curve, that the probability density function $\mu : [0,1] \rightarrow \Re$ has minimum value $\mu_{min} > 0$, and that $\tau_1,\ldots,\tau_m \iid \mu$.  Consider either of the following inference problems:
\begin{enumerate}

\item ($1$-sample)
Fix $\tau_0 \in [0,1]$ and $p_0 = \psi(\tau_0) \in {\mathcal H}$.
Let $h^*$ be a probability distribution on ${\mathcal H}$ with Fr\'{e}chet mean $p^*$.  Fix $s$, and observe $p_1,\ldots,p_s \iid h^*$.  Test the null hypothesis $H_0: p^*=p_0$ at significance level $\alpha$ using the restricted test statistic $\hat{T}_1$ in Section~\ref{restricted.learnt}, embedding as in Theorem~\ref{thm:converge.ED}.

\item ($2$-sample)
Let $h_1^*$ and $h_2^*$ be probability distributions on ${\mathcal H}$ with Fr\'{e}chet means $p_1^*$ and $p_2^*$.  Fix $s_1$ and $s_2$,
 and observe $p_1,\ldots,p_{s_1} \iid h_1^*$ and $p_{s_1+1},\ldots,p_{s_1+s_2} \iid h_2^*$.  Test the null hypothesis $H_0: p_1^*=p_2^*$ at significance level $\alpha$ using the restricted test statistic $\hat{T}_1$ in Section~\ref{restricted2.learnt}, embedding as in Theorem~\ref{thm:converge.ED}.

\end{enumerate}
Then
\[
\lim_{m \rightarrow \infty} \hat{\pi}_1(\tau;m) = \pi_1(\tau;m),
\]
i.e., the power of the restricted test based on the learnt submanifold converges to the power of the restricted test based on the true submanifold as one repeatedly samples the submanifold.
\end{theorem}

\subparagraph{Proof}
For $1$-sample inference, let $p_{s+i}=\psi(\tau_i)$ for $i=1,\ldots,m$.  
Let $x_i = p_i$ and $\hat{x}_i = \W_n \hat{X}_{ni}$ for $i=1,\ldots,s+m$.
Let $n=s+m$ and note that $s$ is fixed.

It follows from Theorem~\ref{thm:2inf} that there exist sequences $n_K \rightarrow \infty$ and $\delta_K \rightarrow 0$ for which
\[
\lim_{K \rightarrow \infty} P \left(
\max_{i=1,\ldots,n_K} \left\|
\hat{x}_i - x_i \right\| \leq \delta_K 
\right) = 1.
\]
Of course, $n_K \rightarrow \infty$ entails $m_K = n_K-s \rightarrow \infty$.

Suppose that $\pi_K \rightarrow 0$ is a decreasing sequence of error probabilities.  Let $D_K$ denote the event
\[
\left\{ \max_{i=1,\ldots,n_K} \left\|
\hat{x}_i - x_i \right\| \leq \delta_K \right\}
\]
and let $E_K$ denote the event that every $x \in {\mathcal H}$ lies within arc length $\delta_K$ of some $x_j = \{ x_{s+1},\ldots,x_{s+m_K} \}$.
Using Theorem~\ref{thm:2inf} and Lemma~\ref{lm:sampling}, choose $m_K$ large enough that $P(D_K \cap E_K) \geq 1-\pi_K$.

As in Sections \ref{SPDconverge} and \ref{EDconverge}, suppose that $\gamma : [0,L] \rightarrow \Re^k$ parametrizes ${\mathcal M}$ by arc length and set $x_i = \gamma(t_i)$ for $i=0,1,\ldots,n_k$.
Let $V_K = \{ x_0,\hat{x}_1,\ldots,\hat{x}_{n_K} \}$ and
let $d_K$ denote shortest path distance on ${\mathcal G}_K$, the $\lambda_K$-neighborhood graph constructed from $V_K$.  By Theorem~\ref{thm:converge.ED}, there exists $\lambda_K \rightarrow 0$ for which each
$\left| \hat{Z}_i-\hat{Z}_0 \right|$ converges in probability to 
$d_M \left( x_0,x_i \right) = |t_i-t_0|$, 
$i=1,\ldots,s$, as $K \rightarrow \infty$.

Next let
\begin{eqnarray*}
\bar{Z}_s = \frac{1}{s} \sum_{i=1}^s \hat{Z}_i & \mbox{ and } &
\bar{t} = \frac{1}{s} \sum_{i=1}^s \hat{t}_i.
\end{eqnarray*}
Because $s$ is fixed, 
$\left| \bar{Z}_s-\hat{Z}_0 \right|$ converges in probability to
$\left| \bar{t}-t_0 \right|$ as $m_K \rightarrow \infty$.
We thus obtain
\[
\hat{T}_1 \left( \A \right) =
\left| \bar{Z}_s-\hat{Z}_0 \right|
\stackrel{P}{\rightarrow}
\left| \bar{t}-\hat{t}_0 \right| =
\left| \mu_{Fr} \left( p_1^*,\ldots,p_s^* \right)-p_0 \right|
\]
as $K \rightarrow \infty$.

Finally, let $\hat{C}_1(m_K)$ denote the $1-\alpha$ quantile of $\hat{T}_1(\A)$ and let $\hat{C}_1 = \lim_{K \rightarrow \infty} C_1(m_K)$.  
Because $\hat{T}_1(\A)$ and $T_1(\A)$ have the same limiting distributions, $\hat{C}_1=C_1$ and
\begin{eqnarray*}
\lim_{K \rightarrow \infty} \hat{\pi}_1 \left( \tau^*;m_K \right)
 & = & 
\lim_{K \rightarrow \infty} P \left( \hat{T}_1 \left( \A \right) \geq \hat{C}_1 \left( m_K \right) \right) \\
 & = &
\lim_{m \rightarrow \infty} P \left( T_1 \left( \A \right) \geq C_1(m) \right) \\
 & = &
\lim_{m \rightarrow \infty} \pi_1(\tau^*;m).
\end{eqnarray*}

For $2$-sample inference, let $p_{s_1+s_2+i}=\psi(\tau_i)$ for $i=1,\ldots,m$.  
Let $x_i = p_i$ and $\hat{x}_i = \hat{X}_{ni}$ for $i=1,\ldots,s_1+s_2+m$.
Let $n=s_1+s_2+m$ and note that $s_1$ and $s_2$ are fixed.
The rest of the proof follows the argument for $1$-sample inference.
\hfill $\Box$

\section{Application}
\label{app}

To illustrate our methods using real-world data, 
we tested a hypothesis about the connectome (wiring diagram) of the right hemisphere of the {\em Drosophila}\/ larval mushroom body, deduced by Eichler et al.\ \cite{Eichler&etal:2017} using electron microscopy.
This connectome is a simple binary directed graph in which vertices correspond to $213$ neurons and edges correspond to synapses.  Our hypothesis concerns the $n=100$ Kenyon Cell (KC) neurons. 

\begin{figure}[hbt]
\centering
\includegraphics[width=0.65\linewidth]{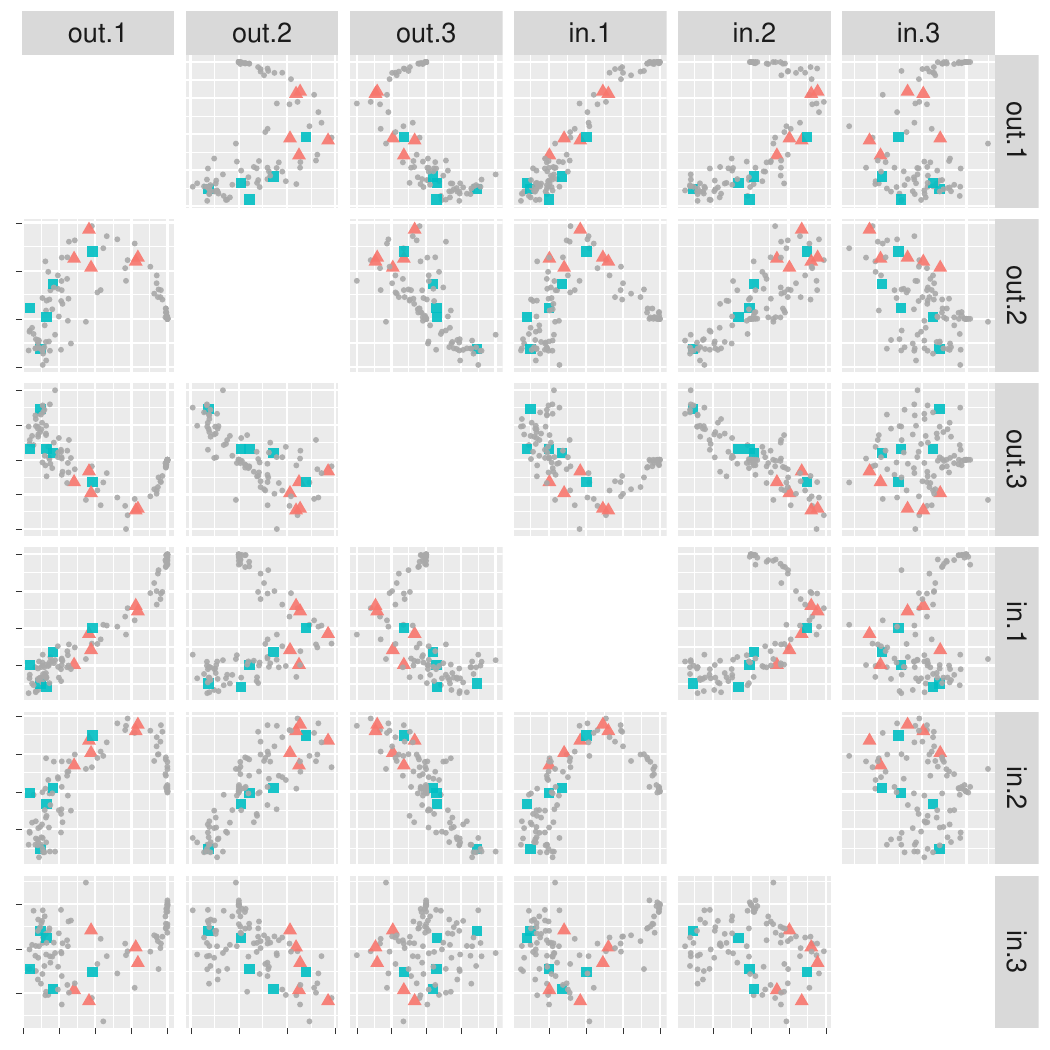}
\caption{
Estimated latent positions of
$n=100$ Kenyon Cell neurons in the right hemisphere larval Drosophila mushroom body connectome,
depicted by the pairs plot of the digraph's $6$-dimensional adjacency spectral embedding representation (three left and three right scaled eigenvectors of the digraph's adjacency matrix).
The possibility that a $1$-dimensional manifold approximates the true latent positions is apparent.
Two samples of $s_i=5$ neurons, identified by distance-to-neuropile, are represented by red triangles and green squares.
}
\label{fig:MWTdrosophila}
\end{figure}

For directed graphs, it is standard practice to construct the adjacency spectral embedding using both the left and the right singular vectors of the adjacency matrix.  Using three of each results in a $6$-dimensional Euclidean representation.  The KC neurons in this representation are displayed in Figure~\ref{fig:MWTdrosophila}.  The plausibility of approximating the latent positions of these neurons by a $1$-dimensional curve is evident.

Let $g(v)$ denote the physical distance in microns of the bundle entry point of KC neuron $v$ from the mushroom body neuropil.  This distance is thought to be a proxy for the age of $v$ \cite{Eichler&etal:2017,drosophila:2017,AA&etal:2018}.
There are $s_1=5$ KC neurons with $g(v)=3689$ (red triangles in Figure~\ref{fig:MWTdrosophila}) and $s_2=5$ KC neurons with $g(v) \in [3811,3928]$ (green squares in Figure~\ref{fig:MWTdrosophila}).  The remaining $m=90$ vertices satisfy either $g(v) \leq 3571$ or $g(v) \geq 4249$.  We regard the $s_1=5$ and $s_2=5$ KC neurons as iid samples from distributions $h_1^*$ and $h_2^*$ on an unknown curve, use the remaining $90$ vertices to facilitate learning the curve, and test the null hypothesis that $h_1^*$ and $h_2^*$ have identical Fr\'{e}chet means.

Neither Section \ref{restricted.learnt} nor \ref{restricted2.learnt} considered practical methods for computing significance probabilities.  For simple null hypotheses, significance probabilities can be estimated by Monte Carlo simulation of $\hat{T}_1$ under the null hypothesis.  For $2$-sample tests, one might perform a permutation test, computing $\hat{T}_1$ for each permutation of $p_1,\ldots,p_{s_1+s_2}$.  Because embedding is not affected by which $p_i$ belongs to which sample, this test is equivalent to computing $\hat{T}_1$ for each permutation of $\hat{Z}_1,\ldots,\hat{Z}_{s_1+s_2}$.  Instead, we performed Welch's approximate $t$-test on $\hat{Z}_1,\ldots,\hat{Z}_{s_1+s_2}$ and compared the resulting significance probability to that resulting from an analogous multivariate test in the ambient $6$-dimensional space.  The univariate test was performed using the \mbox{\tt t.test} function in \mbox{\tt R}; the multivariate test, a variant of the $2$-sample Hotelling $T^2$ test, was performed using 
the \mbox{\tt hotelling.test} function in the \mbox{\tt R} package 
\mbox{\tt Hotelling}.

\subparagraph{Example 1 (continued)}
To explore the implications of using Hotelling's $T^2$ test in the ambient space and univariate $t$ tests for restricted inference, we repeated the analyses described in Example~1, obtaining power estimates of $0.326$ (unrestricted), $0.644$ (restricted to the true submanifold), and $0.846$ (restricted to the learnt submanifold).  Monte Carlo simulation estimated the size of these tests with a nominal significance level of $0.05$ to be 
$0.054$, $0.036$, and $0.047$.  Thus, although these tests have less power than the tests proposed in Section~\ref{test1}, the qualitative characteristics of the results are the same.  In particular, the restricted tests are dramatically more powerful than the unrestricted test.
\hfill $\Box$

\bigskip

To establish the validity of these procedures for the KC data, we first tested equality of Fr\'{e}chet means for each of the ${10 \choose 5} = 252$ ways of partitioning the $2s_i=10$ neurons of interest into two samples of size $s_i=5$. 
Both the multivariate and the univariate tests yielded approximately uniform significance probabilities.
For the observed two samples,
the multivariate Hotelling test in $6$~dimensions yielded a significance probability of $0.275$ and the univariate $t$-test on the $1$-dimensional learnt manifold yielded a significance probability of just $0.032$.  The reduction in significance probability is substantial, demonstrating the value of identifying and exploiting low-dimensional manifold structure for subsequent inference.

\section{Discussion}
\label{discuss}

This investigation continues an ongoing study of restricted inference \cite{mwt:lrt,mwt:ait}.  In \cite{mwt:ait} we considered the setting of statistical submanifolds, i.e., submanifolds that correspond to restricted parametric families of probability distributions.  In the present setting, the submanifolds of interest are curves on which lie latent positions of random dot product graphs.  Our fundamental message is that, even when these curves are unknown, one can learn them well enough to obtain benefits from restricted inference.

To explore the benefits of manifold learning for subsequent inference on random dot product graphs, we have studied $1$- and $2$-sample tests of null hypotheses Fr\'{e}chet means, a natural notion of centrality on a Riemannian manifold.  
The methods that we have proposed and analyzed rely on Isomap to learn the unknown manifold.
One might contemplate the use of other manifold learning procedures, but Isomap is especially well-suited to the inference task we have considered.
In this investigation, the manifolds are curves.  If $\gamma : [0,L] \rightarrow \Re^k$ parametrizes the curve by arc length, then the sample Fr\'{e}chet mean of $\gamma(t_1),\ldots,\gamma(t_s)$ is $\gamma(\bar{t})$, where $\bar{t}$ is the sample mean of $t_1,\ldots,t_s$.
The manifold that Isomap learns is automatically parametrized by arc length, so computing sample means and Euclidean distances in the learnt manifold is inherently analogous to computing Fr\'{e}chet means and geodesic distances on the original curves.
This correspondence underpins our convergence and power analyses.

An exciting by-product of our investigation is a novel convergence analysis of Isomap.  In contrast to \cite{Bernstein&etal:2000}, we consider the application of Isomap to data that do not lie on the manifold of interest.  To ensure convergence, we require that the data converge to the manifold as more data is collected.  Such an assumption might seem fanciful if imposed arbitrarily, but it is automatically satisfied when we use adjacency spectral embedding to estimate the latent positions of random dot product graphs.

The power analysis in Section~\ref{power} demonstrates that, as the number of auxiliary vertices increases, the power of the restricted test that relies on the learnt manifold ($\hat{T}_1$) tends to the power of the test that relies on the true manifold ($T_1$).  Despite the effort required to demonstrate it, the result itself did not surprise us.  What did surprise us were the results of the simulation study reported in Example~1, in which $\hat{T}_1$ outperformed $T_1$.  We do not fully understand this phenomenon, but we note that $\hat{T}_1$ and $T_1$ differ not only with respect to geodesic distance (learnt versus true), but also with respect to point estimates (embedding versus minimum distance).  In effect,
$\hat{T}_1$ fits a curve to the estimated latent positions, whereas $T_1$ forces the true curve to fit the estimated latent positions.  Although adjacency spectral embedding is consistent, the estimated latent positions that it produces are biased for finite sample sizes.  Accordingly, our interpretation of Example~1 is that using learnt distances on a more faithful representation of the data may be better than using true distances on a less faithful representation of the data.  This phenomenon warrants further investigation.

Several technical difficulties also remain to be addressed in future work.  The choice of ambient dimension for adjacency spectral embedding is a general problem, not specific to our present concern with restricted inference.  The construction of the graph used by Isomap to learn the unknown submanifold requires specification of a localization parameter, a problem that is ubiquitous in manifold learning.  Qualitatively, our convergence analysis provides some guidance for choosing this parameter: as the number of latent positions increases, neighborhood size should decrease more slowly than the latent positions fill the submanifold.  However, a specific rule that permits automatic implementation of our methods awaits future development.

Most importantly, this investigation has been concerned entirely with learning $1$-dimensional submanifolds, i.e., curves.  Although the concepts and techniques that we have employed, e.g., Fr\'{e}chet means and Isomap, extend to $d$ dimensions, that case is considerably more challenging.  Key elements of our analysis exploit the fact that curves are locally isometric to Euclidean space.  That is not true of most surfaces, in which case the representations learned by Isomap can never be completely faithful to the actual submanifolds of interest.  Future work on $d$-dimensional submanifolds will have to address difficulties that do not exist in the $1$-dimensional setting.

\section*{Acknowledgments}
This work was partially supported by the Naval Engineering Education Consortium (NEEC), Office of Naval Research (ONR) Award Number N00174-19-1-0011.  The numerical results reported in Example~1 and Section~\ref{app} were obtained by Youngser Park.

\bibliography{$HOME/lib/tex/stat,$HOME/lib/tex/mds,$HOME/lib/tex/math,$HOME/lib/tex/mwt,$HOME/lib/tex/cep,$HOME/lib/tex/bio,$HOME/lib/tex/net}

\end{document}